\documentclass[letterpaper, 10 pt, conference]{ieeeconf}
\IEEEoverridecommandlockouts
\usepackage{cite}
\usepackage{amsmath,amssymb,amsfonts}
\usepackage{algorithmic}
\usepackage{graphicx}
\makeatletter
\let\NAT@parse\undefined
\makeatother
\usepackage[hidelinks,bookmarks=true]{hyperref}
\usepackage[capitalize,nameinlink]{cleveref}
\usepackage{textcomp}
\usepackage[table]{xcolor}
\usepackage{siunitx}
\def\BibTeX{{\rm B\kern-.05em{\sc i\kern-.025em b}\kern-.08em
    T\kern-.1667em\lower.7ex\hbox{E}\kern-.125emX}}
\usepackage{comment}
\usepackage{censor}
\usepackage[english]{babel}
\usepackage[utf8]{inputenc}
\usepackage[acronym, toc]{glossaries}
\graphicspath{{images/}}
\usepackage{wrapfig}
\usepackage{subcaption}
\usepackage{url}
\usepackage[colorinlistoftodos]{todonotes}
\usepackage[T1]{fontenc}
\usepackage{pifont}
\newcommand{\cmark}{\ding{51}}%
\usepackage{subfiles}
\usepackage{csquotes}
\usepackage{adjustbox}
\usepackage{booktabs}
\usepackage{xcolor}
\usepackage{colortbl}
\usepackage{makecell}
\usepackage{multirow}
\usepackage{multicol}

\usepackage{array}
\usepackage{tikz}
\usetikzlibrary{shapes.geometric, positioning, arrows.meta}
\tikzstyle{startstop} = [rectangle, minimum width=25mm, minimum height=8mm, rounded corners,text centered, draw=black, fill=green!10]
\tikzstyle{decision} = [diamond, text centered, draw=black]
\tikzstyle{process} = [ellipse, minimum width=25mm, minimum height=8mm,  text centered, draw=black, fill=orange!10]
\tikzstyle{subtree} = [trapezium, trapezium left angle=70, trapezium right angle=110, minimum width=25mm, minimum height=8mm, text centered, draw=black, fill=green!10]
\tikzstyle{arrow} = [thick,->,>=stealth]
\usepackage{varwidth}

\DeclareSIUnit{\bitunit}{bit}
\DeclareSIUnit{\pixelunit}{px}

\newcommand{\Title}{MultiFly: A Real-World Multimodal Aerial Dataset with\\Annotation-Efficient Label Transfer and Cross-Modal Semantic Consistency}
\title{\Title}

\author{Markus Gross\textsuperscript{1,2,3,4,*}, Andreas Greiner\textsuperscript{1,5,*}, Taehyoung Kim\textsuperscript{1,*}, Sivasubiramaniam Subbiah\textsuperscript{1},\\Tomaž Cotič\textsuperscript{1}, Conrad Christoph\textsuperscript{1}, Oussema Dhaouadi\textsuperscript{2,3,4}, Simon Zieher\textsuperscript{1}, Surya V. Kumar\textsuperscript{1},\\Gordon Elger\textsuperscript{5}, Henri Meeß\textsuperscript{1}, Olaf Wysocki\textsuperscript{4}, Paul Spannaus\textsuperscript{1,$\dagger$}, Daniel Cremers\textsuperscript{2,3}
\thanks{$^{1}$Autonomous Aerial Systems, Fraunhofer Institute IVI}%
\thanks{$^{2}$Computer Vision Group, Technical University of Munich}%
\thanks{$^{3}$Munich Center for Machine Learning (MCML)}%
\thanks{$^{4}$Computer Vision for Digital Twins, University of Cambridge}%
\thanks{$^{5}$Institute of Innovative Mobility, Univ. of Applied Sciences Ingolstadt}%
\thanks{$^{*}$Equal contribution. $^{\dagger}$Correspondence: paul.spannaus@ivi.fraunhofer.de}%
\thanks{{}}%
\thanks{\textit{Preprint.}}%
}

\usepackage{arydshln}

\begin{document}

\makeatletter
  \let\@oldmaketitle\@maketitle% Store \@maketitle
  \renewcommand{\@maketitle}{\@oldmaketitle
  \bigskip
  \centering
    \includegraphics[width=0.85\textwidth]{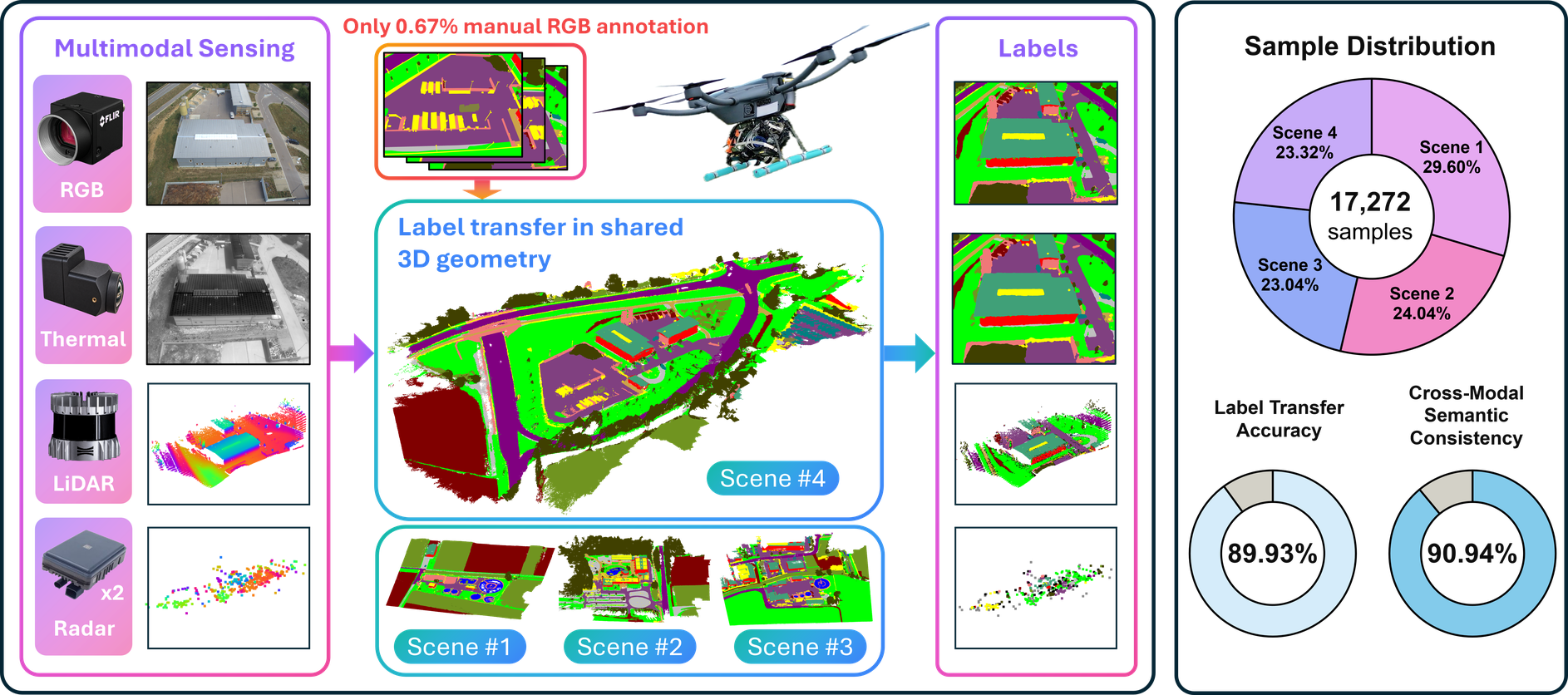}
    \captionof{figure}{Overview of MultiFly. A synchronized and calibrated UAV platform captures multimodal RGB, thermal, LiDAR, and radar data, while our geometry-driven pipeline propagates \SI{0.67}{\%} RGB-only annotations to all four modalities, generating $17,272$ cross-modally consistent semantic labels for four scenes, two altitudes, and 15 semantic categories.}
    \label{fig_banner}
    \vspace{-3mm}
  }

\maketitle
\thispagestyle{empty}
\pagestyle{empty}

\makeatother
\maketitle

\thispagestyle{empty}
\pagestyle{empty}
\setcounter{figure}{1}

\begin{abstract}

We introduce MultiFly, a real-world, low-altitude UAV dataset for semantic perception across RGB, thermal, LiDAR, and radar modalities.
MultiFly provides 17,272 synchronized samples from four suburban scenes with frame-wise annotations for 15 semantic classes, together with calibration and GNSS-RTK/IMU measurements.
To avoid costly and inconsistent modality-specific annotation, we propagate labels from only 115 manually annotated RGB images through shared geometric representations to all four modalities.
This approach generates semantic labels for 17,157 additional RGB images, 17,272 thermal images, 840M LiDAR points, and 3.4M radar points.
Transferred annotations achieve 89.93\% average agreement with held-out manual annotations, and 90.94\% average semantic consistency across all six modality pairs.
We further establish semantic segmentation benchmarks for all four modalities, revealing distinct architectural behavior for dense LiDAR and sparse radar data.
Taken together, MultiFly provides a scalable foundation for multimodal aerial perception and, to the best of our knowledge, the first public real-world low-altitude aerial benchmark that combines consistent frame-wise semantic annotations for RGB, thermal, LiDAR, and radar.
Data at 
{\fontseries{b}\selectfont
\href{https://github.com/markus-42/multifly}{%
    \textcolor{magenta}{\url{https://github.com/markus-42/multifly}}%
}}.

\end{abstract}
\begin{table*}[t]
\centering
\caption{Comparison of real-world multimodal UAV datasets with frame-wise semantic labels (see \cref{sec_related_work}).}
\label{tab_datasets}
\setlength{\aboverulesep}{0pt}
\setlength{\belowrulesep}{0pt}
\renewcommand{\arraystretch}{1.15}
\setlength{\extrarowheight}{1.5pt}
\resizebox{\textwidth}{!}{%
\begin{tabular}{l|cccc|cc|ccc|ccc}
\toprule
\multirow{2}{*}[-0.6ex]{Dataset} 
& \multicolumn{4}{c|}{Perception Modalities} 
& \multicolumn{2}{c|}{Auxiliary Signals} 
& \multirow{2}{*}[-0.6ex]{\shortstack[c]{Sample\\Count}} 
& \multirow{2}{*}[-0.6ex]{Environments} 
& \multirow{2}{*}[-0.6ex]{\shortstack[c]{Altitude\\{[m]}}} 
& \multirow{2}{*}[-0.6ex]{\shortstack[c]{Semantic\\Classes}} 
& \multirow{2}{*}[-0.6ex]{\shortstack[c]{Cross-Modal\\Semantic Consistency}}
& \multirow{2}{*}[-0.6ex]{\shortstack[c]{Manual Correction\\Not Required}} \\
\cmidrule(lr){2-5} \cmidrule(lr){6-7}
& RGB & Thermal & LiDAR & Radar & IMU & GNSS & & & & & \\
\midrule
IndraEye~\cite{manjunath2025indraeye}
& \cmark & \cmark & - & - & - & - 
& $5{,}612$ 
& urban 
& $7$--$30$ 
& $13$ 
& -
& - \\
\rowcolor[HTML]{E6E6E6} 
MVUAV~\cite{cheng2025mvuav}
& \cmark & \cmark & - & - & - & - 
& $2{,}183$ 
& urban 
& $5$--$20$ 
& $36$ 
& -
& - \\
CART~\cite{lee2024cart}
& \cmark & \cmark & - & - & - & - 
& $2{,}282$ 
& rural 
& $40$ 
& $11$ 
& -
& - \\
\rowcolor[HTML]{E6E6E6} 
Kust4K~\cite{ouyang2025kust4k}
& \cmark & \cmark & - & - & - & - 
& $4{,}024$ 
& urban 
& - 
& $8$ 
& -
& - \\
SegFly~\cite{gross2026segfly}
& \cmark & \cmark & - & - & - & - 
& $15{,}007$ 
& urban, industrial, rural 
& $30$, $40$, $50$ 
& $15$ 
& \cmark
& \cmark \\
\rowcolor[HTML]{E6E6E6} 
UAVScenes~\cite{wang2025uavscenes}
& \cmark & - & \cmark & - & - & \cmark 
& $120{,}000$ 
& rural, airport 
& $80$, $90$, $130$ 
& $18$ 
& \cmark
& - \\
\midrule
MultiFly (ours)
& \cmark & \cmark & \cmark & \cmark & \cmark & \cmark 
& $17{,}272$ 
& suburban 
& $30$, $50$ 
& $15$ 
& \cmark
& \cmark \\
\bottomrule
\end{tabular}%
}
\end{table*}

\section{Introduction}
\label{sec_intro}

Semantic scene understanding is fundamental to aerial perception~\cite{gross2026safeland, nguyen2019arialSurveillanceInspection, bultmann2023aerialDisasterRescue}. However, unmanned aerial vehicle (UAV) data remains substantially underrepresented in large-scale multimodal datasets, which increasingly drive modern foundation models. Notably, this is not simply a data-volume problem: UAV perception introduces distinct scene statistics, object scales, and viewing geometries, ranging from nadir to oblique perspectives, that are poorly represented in conventional ground-based datasets. Autonomous driving illustrates the contrast, where large-scale multimodal datasets have enabled systematic alignment across sensors, while similarly comprehensive resources for UAV perception remain limited.

This gap is particularly evident for semantic segmentation. Existing aerial datasets generally provide a single modality or limited modality combinations~\cite{nigam2018aeroscapes,gross2026segfly,wang2025uavscenes}, while no public real-world UAV dataset, to the best of our knowledge, provides synchronized frame-wise semantic annotations across RGB, thermal, LiDAR, and radar. Moreover, these modalities differ substantially in geometry and sampling, making consistent annotation difficult and costly.

To address these challenges, we introduce MultiFly, a calibrated and time-synchronized four-modal aerial sensing platform and semantic dataset.
Our annotation framework uses 3D scene geometry as a common semantic interface: a~sparse set of manually annotated RGB images is lifted into 3D and transferred to synchronized RGB, thermal, LiDAR, and radar measurements.
This yields dense multimodal annotations from only 115 manually labeled RGB images, while maintaining a common taxonomy and geometric reference.

We summarize our main contributions as follows:
\begin{itemize}
    \item \textbf{MultiFly dataset release:}
    \(17{,}272\) synchronized RGB, thermal, LiDAR, and radar samples across $4$ suburban scenes, covering \(56{,}700\,\mathrm{m}^{2}\) and $15$ semantic classes, with full sensor calibration and state information.

    \item \textbf{Annotation-efficient label transfer:}
    Semantic labels are propagated from only 115 manually annotated RGB images to additional \(17{,}157\) RGB frames, \(17{,}272\) thermal frames, \(8.4 \times 10^{8}\) LiDAR points, and \(3.4 \times 10^{6}\) radar points, achieving \(89.93\%\) average agreement with held-out manual annotations, and \(90.94\%\) average semantic consistency across all six modality pairs.

    \item \textbf{Benchmarking:}
    We establish semantic segmentation benchmarks on MultiFly for RGB, thermal, LiDAR, and radar modalities, and analyze architectural influences between dense LiDAR and sparse radar segmentation.
\end{itemize}

\section{Related Work}
\label{sec_related_work}

\subsection{Multimodal Semantic Aerial Datasets}
Semantic perception has evolved from single-sensor benchmarks~\cite{brostow_2008_camvid, munoz2009oakland3d} toward multimodal datasets that enable learning across complementary sensors. In autonomous driving, early resources established core camera-LiDAR perception tasks, while second-generation datasets such as nuScenes~\cite{caesar2020nuscenes} expanded to large-scale, synchronized multimodal sensing. This shift enabled not only sensor fusion, but also temporal and multi-task perception, prediction, and planning, and increasingly multimodal foundation models that unify these capabilities.

UAV perception has yet to undergo a comparable transition. Existing aerial datasets remain fragmented across sensing modalities and annotation formats: most are unimodal~\cite{arroyo2024radarUnimodalDataset, lee2026lidarUnimodalDataset, nigam2018aeroscapes}, while others combine only two modalities~\cite{lee2024cart, li2024mars-lvig, wang2025uavscenes}, rely on synthetic data~\cite{rizzoli2023syndrone, zou2025uav-mm3d}, or provide map-wise rather than frame-wise annotations~\cite{koelle2021hessigheim3d, lin2022urbanScene3d}. Consequently, the data required to develop and systematically evaluate unified multimodal UAV perception models remains largely unavailable. In particular, no public real-world low-altitude UAV dataset, to the best of our knowledge, provides synchronized frame-wise semantics across RGB, thermal, LiDAR, and radar.
MultiFly provides this missing substrate (Table~\ref{tab_datasets}), combining synchronized and registered RGB, thermal, LiDAR, and radar data with inertial measurements from Inertial Measurement Units (IMUs) and global positioning from a Global Navigation Satellite System (GNSS) with real-time kinematic (RTK) corrections.

\subsection{Cross-Modal Semantic Consistency}
Beyond sensor coverage, multimodal learning requires semantic supervision that is consistent across modalities. Independent annotation can introduce conflicting boundaries and labels, motivating geometry-based approaches that propagate semantics through a shared representation. KITTI-360~\cite{liao2023kitti-360} transfers labels between images and 3D geometry, while WildScenes~\cite{vid2024wildScenes} projects multi-view image annotations into LiDAR.
UAVScenes~\cite{wang2025uavscenes} similarly annotates reconstructed 3D maps and renders them into camera frames before projecting labels onto LiDAR, while SegFly~\cite{gross2026segfly} adopts a 2D-3D-2D pipeline to lift sparse RGB annotations into a reconstructed point cloud and renders them into RGB and thermal images. These approaches demonstrate the feasibility of geometry-driven annotation, but remain limited in modality coverage, and require manual verification and correction.

MultiFly extends this paradigm to four modalities through a unified geometry-driven pipeline, propagating sparse RGB annotations to RGB, thermal, LiDAR, and radar, while ensuring high semantic consistency without the necessity of manual correction. It therefore offers not only broader sensor coverage, but a shared frame-wise semantic reference for multimodal UAV perception and cross-modal evaluation.

\section{MultiFly Platform}
\label{sec_multifly_platform}

\begin{table*}[t]
\centering
\caption{Sensor specifications of the MultiFly sensor suite. For more details, see \cref{sec_multifly_platform}.}
\label{tab_sensors}
\setlength{\aboverulesep}{0pt}
\setlength{\belowrulesep}{0pt}
\renewcommand{\arraystretch}{1.15}
\setlength{\extrarowheight}{1.5pt}
\resizebox{\textwidth}{!}{%
\begin{tabular}{c|cccc|ccc|cccc} 
\toprule
\multirow{2}{*}[-0.6ex]{Sensor} &
\multirow{2}{*}[-0.6ex]{Model} &
\multirow{2}{*}[-0.6ex]{Count} &
\multirow{2}{*}[-0.6ex]{Frequency} &
\multirow{2}{*}[-0.6ex]{Data Type} &
\multicolumn{3}{c|}{Field of View} &
&
\multicolumn{3}{c}{Resolution}
\\
\cmidrule(lr){6-8} \cmidrule(lr){10-12}
&
&
&
&
&
Range &
Azimuth &
Elevation &
&
Range &
Azimuth &
Elevation
\\
\midrule
\begin{tabular}[c]{@{}c@{}}RGB Camera\\and Lens\end{tabular} &
\begin{tabular}[c]{@{}c@{}}FLIR Blackfly S GigE\\BFS-PGE-31S4C-C\\with Tamron M112FM06\end{tabular} &
$1$ &
\SI{10}{Hz} &
\begin{tabular}[c]{@{}c@{}}\SI[number-unit-product={-}]{8}{\bitunit} Bayer\\image\end{tabular} &
- &
\SI{62}{\degree} &
\SI{48}{\degree} &
&
- &
\SI{2048}{\pixelunit} &
\SI{1536}{\pixelunit}
\\
\rowcolor[HTML]{E6E6E6} 
Thermal Camera &
\begin{tabular}[c]{@{}c@{}}IRCAM NIK III\\microbolometer camera\end{tabular} &
$1$ &
\begin{tabular}[c]{@{}c@{}}\SI{20}{Hz} raw\\\SI{10}{Hz} selected\end{tabular} &
\begin{tabular}[c]{@{}c@{}}\SI[number-unit-product={-}]{16}{\bitunit} monochrome\\thermal image\end{tabular} &
- &
\begin{tabular}[c]{@{}c@{}}\SI{57.5}{\degree}\end{tabular} &
\begin{tabular}[c]{@{}c@{}}\SI{47.39}{\degree}\end{tabular} &
&
- &
\SI{1280}{\pixelunit} &
\SI{1024}{\pixelunit}
\\
\rowcolor[HTML]{FFFFFF} 
LiDAR &
Ouster OS1-128 &
$1$ &
\SI{10}{Hz} &
\begin{tabular}[c]{@{}c@{}}x, y, z, intensity,\\ambient, reflectivity\end{tabular} &
\begin{tabular}[c]{@{}c@{}}\SI{90}{m} recommended\\\SI{200}{m} maximum\end{tabular} &
\begin{tabular}[c]{@{}c@{}}\SI{360}{\degree}\\\SI{90}{\degree} selected\end{tabular} &
\SI{45}{\degree} &
&
- &
\begin{tabular}[c]{@{}c@{}}$2048$ cols\\per scan\end{tabular} &
\begin{tabular}[c]{@{}c@{}}$128$\\channels\end{tabular}
\\
\rowcolor[HTML]{E6E6E6} 
Radar &
Continental ARS 548 RDI &
$2$ &
\SI{10}{Hz} &
\begin{tabular}[c]{@{}c@{}}x, y, z, velocity, RCS,\\range, azimuth, elevation\end{tabular} &
\begin{tabular}[c]{@{}c@{}}\SI{301}{m}\\\SI{1514}{m} extended\end{tabular} &
\begin{tabular}[c]{@{}c@{}}\SI{100}{\degree}\\\SI{120}{\degree} processed\end{tabular} &
\begin{tabular}[c]{@{}c@{}}\SI{8}{\degree}\\\SI{28}{\degree} below \SI{100}{m}\end{tabular} &
&
\SI{0.22}{m} &
\begin{tabular}[c]{@{}c@{}}\SI{1.2}{\degree} near boresight\\\SI{1.68}{\degree} at \SI{\pm45}{\degree}\end{tabular} &
\SI{2.3}{\degree}
\\
\rowcolor[HTML]{FFFFFF}
GNSS-RTK / INS &
\begin{tabular}[c]{@{}c@{}}OxTS xRED\\with dual GNSS receivers\end{tabular} &
$1$ &
\SI{100}{Hz} &
\begin{tabular}[c]{@{}c@{}}time, latitude, longitude, altitude,\\linear velocity, angular velocity,\\yaw, pitch, roll, status\end{tabular} &
- &
- &
- &
&
- &
- &
-
\\
\bottomrule
\end{tabular}%
}
\vspace{-3mm}
\end{table*}

\subsection{Design}
\label{sec_design}

\begin{figure}[t]
    \centering
    \includegraphics[width=0.58\columnwidth]{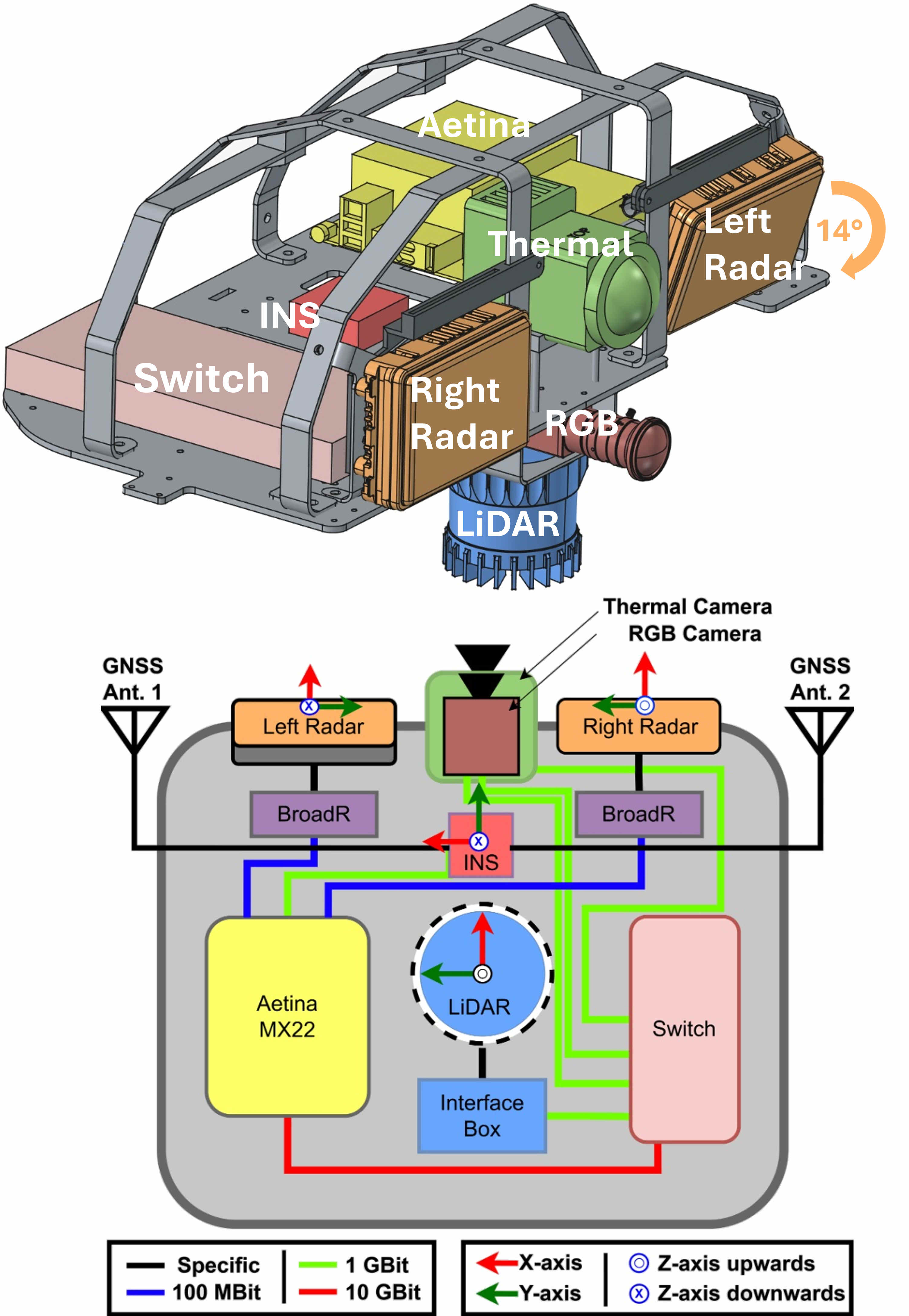}
    \caption{MultiFly sensor rig (see \cref{sec_design}).}
    \label{fig_multifly_platform}
\end{figure}

To acquire multimodal aerial data, we propose the MultiFly platform shown in \cref{fig_multifly_platform}, which integrates one forward-facing RGB camera, one thermal camera, one spinning LiDAR, two forward-facing radars, and a GNSS-RTK-aided inertial navigation system (INS), with detailed specifications listed in \cref{tab_sensors}.
One radar is pitched downward by $14^\circ$ to increase the effective vertical field of view (FOV), while the whole sensor rig is pitched downward by $45^\circ$ to jointly observe the flight direction and ground below the UAV.
All perception sensors are calibrated with respect to the LiDAR reference frame and synchronized to a common time base, while the INS provides the platform pose.
Perception data are recorded at $10\,\mathrm{Hz}$ on an Aetina AIB MX22 edge computing unit using Docker and ROS 2 Humble, together with dedicated onboard power (DOGCOM 10000 mAh Li-Ion), networking (NetGear GSM110MX switch), and storage hardware.
Example synchronized RGB, thermal, LiDAR, and radar measurements are shown in \cref{fig_multifly_sample}.

\begin{figure}[t]
    \centering
    \includegraphics[width=0.6\columnwidth]{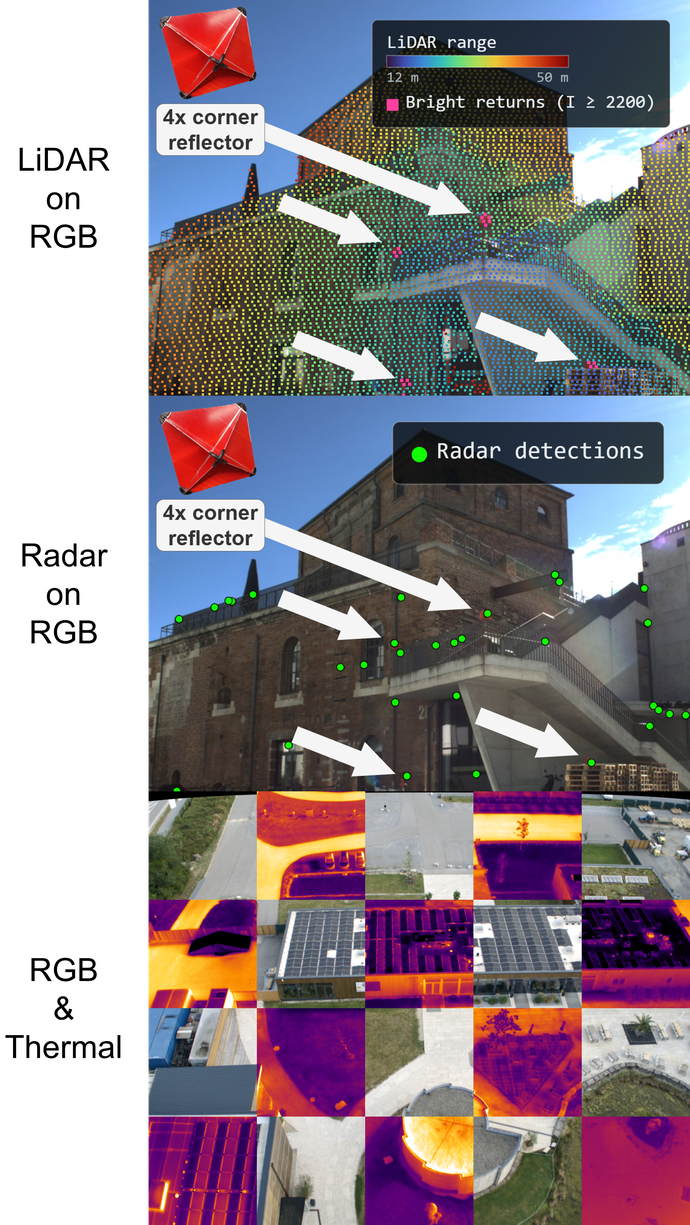}
    \caption{Registration results (see \cref{subsec_registration}).}
    \label{fig_registration_results}
\end{figure}

\subsection{Time Synchronization}
\label{sec_sync}

All sensors are synchronized to a common GNSS-disciplined time base using the Precision Time Protocol (PTP).
The INS acts as the PTP grandmaster, while the Aetina computer operates as a boundary clock and distributes the time reference to the sensor network.
While PTP establishes a common time base, sensor-specific triggering and phase compensation align the actual acquisition times across modalities.
More specifically, sensor acquisition is aligned with the LiDAR's \SI{10}{\hertz} scan cycle, corresponding to a fixed rotation period of \SI{100}{\milli\second}.
The rotating LiDAR generates a hardware trigger for the forward-facing RGB camera, with a configurable delay accounting for the exposure time and the camera's orientation relative to the LiDAR scan, effectively aligning the acquisition with the LiDAR beam passing through the camera FOV.
The radars operate at the same nominal frequency, but their measurement epochs have a deterministic phase offset relative to the LiDAR cycle, which
is compensated using the driver's \texttt{cycle\_offset} parameter, resulting in radar--LiDAR deviations within \SI{\pm3}{\milli\second}.
Since the thermal camera does not support hardware triggering, its acquisition is controlled using timestamped GigE Vision Scheduled Action Commands, which are executed according to its PTP-synchronized clock.

\subsection{Sensor Calibration and Registration}
\label{subsec_sensor_calibration}

\textbf{Extrinsics.}
We organize the sensor extrinsics as a calibration tree rooted at the LiDAR, which serves as the common reference frame. The tree consists of three edges:

\paragraph{RGB--LiDAR} From a static recording of a structured scene, we manually select natural landmarks that are sharp in both modalities and estimate the rigid transform by solving a perspective-$n$-point problem.

\paragraph{Radar--LiDAR} Corner reflectors are placed in the shared sensor FOV. Corresponding radar and LiDAR detections are paired and the rigid transform is recovered in closed form via Singular Value Decomposition.

\paragraph{RGB--Thermal} The two cameras are stereo-calibrated on checkerboard corners observed by both sensors.
Composing the RGB--Thermal and RGB--LiDAR transforms places the thermal camera in the LiDAR frame.

\textbf{Intrinsics.}
We calibrate the RGB camera and the thermal camera independently with Zhang's
planar method~\cite{zhang2000camCalibration} from multiple checkerboard views.
The RGB target is a conventional checkerboard, while the thermal target is a steel plate
carrying plotter-cut vinyl squares, whose emissivity contrast against the exposed
steel renders the corners detectable in the thermal band.
LiDAR and radar use their factory intrinsics.

\subsection{Registration.}
\label{subsec_registration}

Radar detections are transformed into the LiDAR frame, and LiDAR-frame measurements are projected into either image using the calibrated extrinsics, intrinsics, and distortion models.

Furthermore, RGB--thermal registration requires scene geometry because the nonzero camera baseline induces depth-dependent parallax. Consequently, the calibrated rigid extrinsic alone does not define a dense pixel-to-pixel mapping.
In principle, the synchronized and calibrated LiDAR measurements could provide the required depth for reprojecting RGB observations into the thermal view. 
Instead, we follow SegFly~\cite{gross2026segfly} and use the metric RGB-based 3D reconstruction required by the label-transfer pipeline (\cref{sec_rgb_and_thermal_label_transfer}).
This reconstruction already provides scene-wide geometry and
RGB image--point correspondences.
Naturally, using the same geometric representation for both registration and semantic label transfer ensures consistency between both stages.
Our choice therefore does not reflect a limitation of LiDAR, but avoids
introducing a separate geometric representation for registration. Moreover, unlike SegFly~\cite{gross2026segfly}, whose acquisition setting requires an independent thermal reconstruction and iterative closest point (ICP) alignment, our cameras are synchronized and rigidly calibrated.
We therefore retain the fixed RGB--thermal extrinsic and require neither thermal reconstruction nor ICP.

Technically, for each synchronized RGB--thermal pair, the thermal camera pose is obtained by composing the corresponding RGB reconstruction pose with the fixed RGB--thermal extrinsic.
Reconstructed landmarks jointly visible in both cameras are projected into their undistorted image planes using the calibrated camera models.
Because each pair of projections represents the same physical
3D point, it identifies where scene content visible to the thermal camera appears in the corresponding RGB frame.
Following SegFly, these projected landmarks determine the common image support used to warp the original RGB image.
In particular, the RGB image is resampled from its distorted image plane into a common undistorted pinhole domain and subsequently mapped into the thermal distortion space.
The valid overlap is cropped and resized to the thermal resolution, while pixels mapped outside the valid common support are masked.
This process yields a registered RGB image on the pixel grid of every thermal image.

Registration results in \cref{fig_registration_results} exhibit precise cross-modal alignment and high geometric fidelity.

\section{Annotation-Efficient Label Transfer}
\label{sec_cross_modal_semantic_label_transfer}

Our goal is to generate dense semantic annotations for all four sensor modalities from only a sparse set of manually annotated RGB source views. With these, we aim to produce the complete collections of time-synchronized, frame-wise semantic labels \(\widehat{\mathcal{Y}}^{\mathrm{RGB}}\), \(\widehat{\mathcal{Y}}^{\mathrm{Th}}\), \(\widehat{\mathcal{Y}}^{\mathrm{Lid}}\) and \(\widehat{\mathcal{Y}}^{\mathrm{Rad}}\) for the RGB, thermal, LiDAR and radar modalities, respectively.  
The following subsections describe how these four sets are obtained via geometry-driven label transfer.
While the individual transfer operations build on established geometry-driven methods~\cite{gross2026segfly,liao2023kitti-360,vid2024wildScenes}, our contribution is their modality-specific adaptation and integration into a unified aerial framework that derives consistent RGB, thermal, LiDAR, and radar labels from the same sparse set of manually annotated RGB views.

\subsection{RGB and Thermal Semantic Label Transfer}
\label{sec_rgb_and_thermal_label_transfer}
We build on SegFly's geometry-driven 2D--3D--2D label-transfer pipeline~\cite{gross2026segfly} and use the shared RGB reconstruction and calibrated RGB--thermal geometry introduced in \cref{subsec_registration}.

\subsubsection{\textbf{2D-to-3D}}
\label{subsubsec_2d_to_3d}
The classical image-based RGB 3D reconstruction introduced in \cref{subsec_registration} provides the point cloud \(\mathcal{P}^{\mathrm{RGB}}\) and the image--point correspondences.
Manual semantic annotations from a sparse set of spatially stratified RGB source views are then lifted into the point cloud via the image--point correspondences that follow naturally from 3D reconstruction.
Subsequent neighborhood-based completion then yields the RGB-derived semantic point cloud
\begin{equation}
    \mathcal{P}^{\mathrm{RGB}}_{\mathrm{sem}}
    =
    \{(\mathbf{x}^{\mathrm{RGB}}_m,c_m)\}_{m=1}^{M_{\mathrm{RGB}}}.
    \label{eq_rgb_sem_pcl}
\end{equation}

\subsubsection{\textbf{3D-to-2D}}
\label{subsubsec_3d_to_2d}
The semantic point cloud \(\mathcal{P}^{\mathrm{RGB}}_{\mathrm{sem}}\) is then rendered into all remaining RGB images and into the thermal images using SegFly’s visibility-aware Z-buffered rendering and the camera poses described in \cref{subsec_registration}.
This produces complete sets of dense pseudo-label maps
\begin{equation}
    \widehat{\mathcal{Y}}^{\mathrm{RGB}}
    =
    \{\widehat{\mathbf{Y}}^{\mathrm{RGB}}_n\}_{n=1}^{N_{\mathrm{RGB}}}
    \quad\text{and}\quad
    \widehat{\mathcal{Y}}^{\mathrm{Th}}
    =
    \{\widehat{\mathbf{Y}}^{\mathrm{Th}}_n\}_{n=1}^{N_{\mathrm{Th}}}.
    \label{eq_pseudo_label_sets}
\end{equation}
Because registration and semantic rendering use the same metric reconstruction and fixed RGB--thermal extrinsic, the resulting pseudo-labels are geometrically consistent across both modalities.

\subsection{LiDAR and Radar Semantic Label Transfer}
\label{sec_align_lidar_radar}

The manually annotated RGB source views introduced in
\cref{sec_rgb_and_thermal_label_transfer} are selected in a spatially stratified manner and thus
provide broad spatial coverage of each scene despite sparse manual supervision.
This coverage allows their semantic labels to be transferred to the LiDAR and
radar measurements without requiring additional modality-specific manual
annotation.\\

\subsubsection{\textbf{LiDAR Label Transfer}}
\label{subsubsec_lidar_sem}
We process each scene at each altitude independently. The individual LiDAR scans are transformed into the common scene frame and aggregated into the scene-level point cloud
\begin{equation}
\mathcal{P}^{\mathrm{Lid}}=\{\mathbf{x}^{\mathrm{Lid}}_m\}_{m=1}^{M_{\mathrm{Lid}}},
\end{equation}
while retaining the association of every point with its original scan.
To transfer the manually annotated semantics, we render \(\mathcal{P}^{\mathrm{Lid}}\) into each annotated RGB source view using visibility-aware (Z-buffered) rendering and sample the corresponding labels, similar to the strategies of \cite{liao2023kitti-360, vid2024wildScenes}.
A point may receive observations from multiple views. We therefore assign the majority class if its relative frequency exceeds \(\tau_{\mathrm{Lid}}=0.75\). Otherwise, the point is marked as unlabeled. This yields the semantic LiDAR point cloud
\begin{equation}
    \mathcal{P}^{\mathrm{Lid}}_{\mathrm{sem}}
    =
    \{(\mathbf{x}^{\mathrm{Lid}}_m,\widehat{c}^{\mathrm{Lid}}_m)\}_{m=1}^{M_{\mathrm{Lid}}},
    \label{eq_lid_sem_pcl}
\end{equation}
where \(\widehat{c}^{\mathrm{Lid}}_m\in\mathcal{C}\cup\{c_{\mathrm{unl}}\}\).
Using the retained point-to-scan associations, the labels are mapped back to the original scans. This produces the complete set of frame-wise semantic LiDAR annotations
\begin{equation}
    \widehat{\mathcal{Y}}^{\mathrm{Lid}}
    =
    \{\widehat{\mathbf{Y}}^{\mathrm{Lid}}_k\}_{k=1}^{K_{\mathrm{Lid}}},
    \label{eq_lidar_label_set}
\end{equation}
where \(K_{\mathrm{Lid}}\) is the number of LiDAR scans. Points labeled \(c_{\mathrm{unl}}\) are treated as ignore labels during supervised training.\\

\subsubsection{\textbf{Radar Label Transfer}}
\label{subsubsec_radar_sem}
We treat the two radar sensors as a single logical radar for clarity. Analogous to the LiDAR case, the individual radar scans are transformed into the common scene frame and aggregated into a scene-level point cloud
\begin{equation}
\mathcal{P}^{\mathrm{Rad}}=\{\mathbf{x}^{\mathrm{Rad}}_m\}_{m=1}^{M_{\mathrm{Rad}}},
\end{equation}
while retaining the association of every point with its original scan.
Because radar measurements contain a considerable number of ghost targets and other outliers ($18.8\%$ on average in our data), we apply two consecutive pre-processing steps before label transfer.
First, a RANSAC-based plane fit ($\tau=0.2\,\mathrm{m}$) estimates the ground surface.
Points lying more than two standard deviations below the fitted plane are discarded to preserve genuine ground returns while removing clear outliers. Second, DBSCAN clustering (\(\varepsilon=1.5\,\mathrm{m}\), \(\mathtt{min\_points}=25\)) is applied to remove remaining sparse ghost detections.
Points removed by either step are marked as invalid (\(c_{\mathrm{inv}}\)).
After pre-processing, semantic labels are transferred exactly as for LiDAR: the cleaned point cloud is rendered into the annotated RGB source views using visibility-aware (Z-buffered) rendering, and each point is assigned the majority class if its relative frequency exceeds \(\tau_{\mathrm{Rad}}=0.75\). Otherwise, it is marked unlabeled. This yields the semantic radar point cloud
\begin{equation}
    \mathcal{P}^{\mathrm{Rad}}_{\mathrm{sem}}
    =
    \{(\mathbf{x}^{\mathrm{Rad}}_m,\widehat{c}^{\mathrm{Rad}}_m)\}_{m=1}^{M_{\mathrm{Rad}}},
    \label{eq_rad_sem_pcl}
\end{equation}
where \(\widehat{c}^{\mathrm{Rad}}_m\in\mathcal{C}\cup\{c_{\mathrm{unl}},c_{\mathrm{inv}}\}\). Using the retained point-to-scan associations, the labels are mapped back to the original scans, producing the complete set of frame-wise semantic radar annotations
\begin{equation}
    \widehat{\mathcal{Y}}^{\mathrm{Rad}}
    =
    \{\widehat{\mathbf{Y}}^{\mathrm{Rad}}_k\}_{k=1}^{K_{\mathrm{Rad}}}.
    \label{eq_radar_label_set}
\end{equation}
Points labeled \(c_{\mathrm{unl}}\) or \(c_{\mathrm{inv}}\) can be treated as ignore labels during supervised training.

\begin{table}[t]
\centering
\caption{MultiFly dataset statistics (see \cref{subsec_dataset_statistics}).}
\label{tab_multifly_statistics}
\setlength{\aboverulesep}{0pt}
\setlength{\belowrulesep}{0pt}
\renewcommand{\arraystretch}{1.25}
\setlength{\extrarowheight}{1pt}
\setlength{\tabcolsep}{3pt}

\resizebox{\linewidth}{!}{%
\begin{tabular}{l|cc|cc|cc|cc|c}
\toprule
    & \multicolumn{8}{c|}{MultiFly Scenes and Altitudes}
    & \multirow{3}{*}{Total} \\
\cmidrule(lr){2-9}
    & \multicolumn{2}{c|}{\#1}
    & \multicolumn{2}{c|}{\#2}
    & \multicolumn{2}{c|}{\#3}
    & \multicolumn{2}{c|}{\#4}
    & \\
\cmidrule(lr){2-3}
\cmidrule(lr){4-5}
\cmidrule(lr){6-7}
\cmidrule(lr){8-9}
    Statistic & \SI{30}{\meter} & \SI{50}{\meter}
    & \SI{30}{\meter} & \SI{50}{\meter}
    & \SI{30}{\meter} & \SI{50}{\meter}
    & \SI{30}{\meter} & \SI{50}{\meter}
    & \\
\midrule

Synchronized samples
    & 2,209 & 2,903
    & 1,987 & 2,166
    & 1,766 & 2,214
    & 2,097 & 1,930
    & \textbf{17,272} \\

IMU/GNSS-RTK pose
    & $\checkmark$ & $\checkmark$
    & $\checkmark$ & $\checkmark$
    & $\checkmark$ & $\checkmark$
    & $\checkmark$ & $\checkmark$
    & \textemdash \\

Covered area [\si{\square\meter}]
    & \multicolumn{2}{c|}{17,400}
    & \multicolumn{2}{c|}{11,100}
    & \multicolumn{2}{c|}{13,000}
    & \multicolumn{2}{c|}{15,200}
    & \textbf{56,700} \\
\hline

\bottomrule
\end{tabular}%
}
\end{table}

\begin{figure}[t]
    \centering
    \includegraphics[width=\columnwidth]{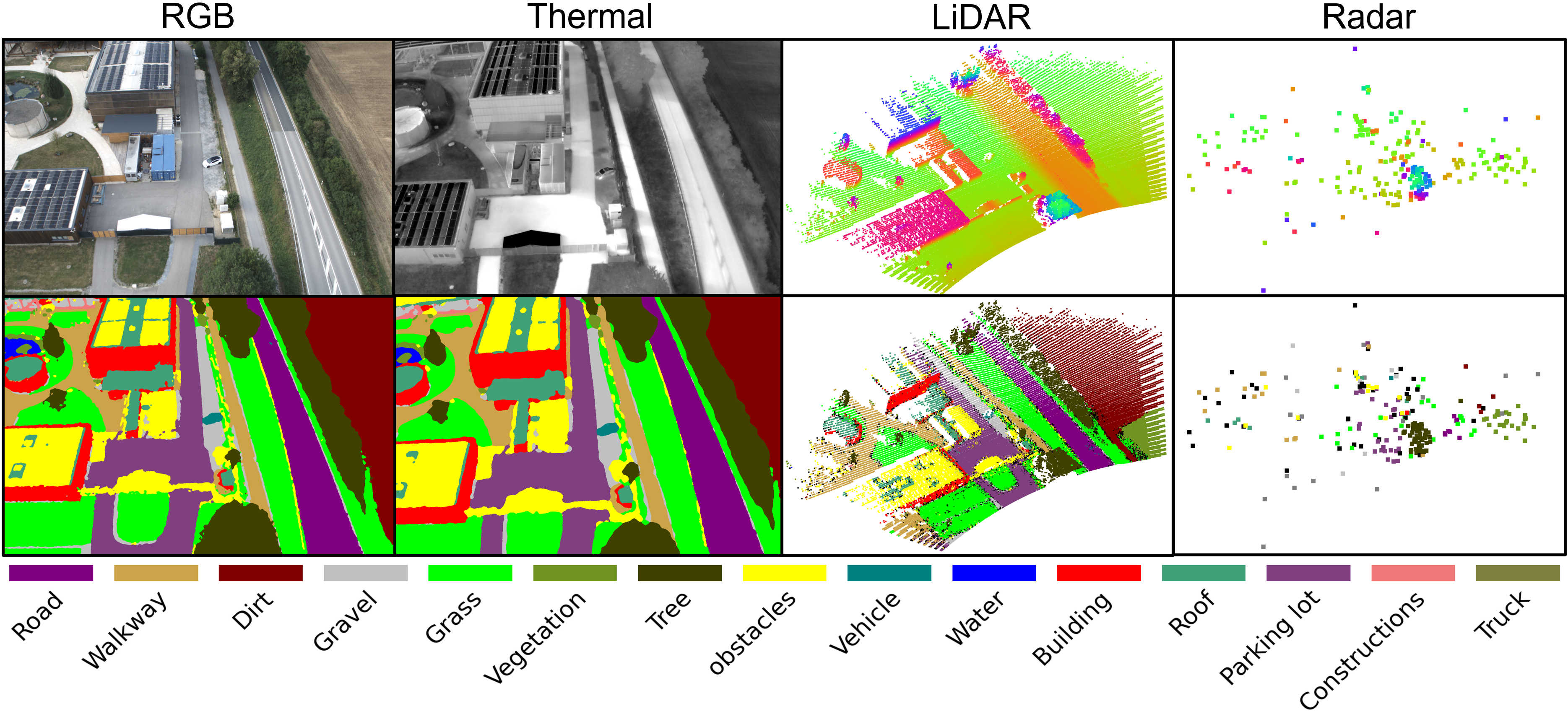}
    \caption{MultiFly sample with semantic classes (Sec.~\ref{subsec_dataset_statistics}).}
    \label{fig_multifly_sample}
\end{figure}

\section{MultiFly Dataset}
\label{sec_dataset}

\subsection{Data Acquisition and Source Annotation}
\label{subsec_data_acquisition_and_source_annotation}
We collect MultiFly using the EmQopter Q6500 UAV (\cref{fig_banner}) together with the calibrated and time-synchronized platform described in \cref{sec_multifly_platform}.
The acquisition campaign comprises four suburban scenes within the metropolitan area of Ingolstadt in Germany, recorded during summer under sunny and cloudy conditions.
Each scene is surveyed at nominal flight altitudes of \(30\,\mathrm{m}\) and \(50\,\mathrm{m}\) above ground level.
At each altitude, the UAV follows an automated double-grid trajectory, in line with the acquisition strategy of OccuFly~\cite{gross2026occufly}.
The flight speed is \(3\,\mathrm{m}\,\mathrm{s}^{-1}\) and the flight lines are planned with a nominal RGB side overlap of \SI{40}{\%}.
During all flights the sensor stack is pitched downward by $45^{\circ}$ relative to the horizontal to observe both the flight direction and the terrain below.
To reduce temporal redundancy while retaining meaningful scene variation, the synchronized 10 Hz perception streams are subsampled by selecting every second frame at \SI{30}{m} and every third frame at \SI{50}{m}.
The INS records at \SI{100}{Hz} and its complete navigation stream is preserved at the original temporal resolution.
The LiDAR records a full $360^{\circ}$ azimuth during acquisition.
During offline preprocessing, each LiDAR scan is restricted to the $90^{\circ}$ sector facing forward and aligned with the RGB camera.

For label propagation, $115$ RGB images are manually annotated using the $15$ semantic classes adopted from SegFly~\cite{gross2026segfly}, and constitute the only manual annotations used for data generation.
No thermal images, LiDAR points, or radar points are manually annotated.

\subsection{Dataset Statistics}
\label{subsec_dataset_statistics}
\Cref{tab_multifly_statistics} summarizes the final MultiFly dataset by scene and acquisition altitude.
MultiFly contains $17{,}272$ synchronized multimodal samples.
Each sample contains one RGB image, one corresponding thermal image, one LiDAR scan restricted to the forward $90^{\circ}$ sector, and one measurement set from each of the two radar sensors.
Every sample is associated with a six degree of freedom pose obtained from the INS.
The complete \SI{100}{Hz} navigation stream is additionally provided at its native temporal resolution, preserving all measurements between consecutive \SI{10}{Hz} perception samples.

MultiFly provides frame-wise semantic annotations in a shared taxonomy of $15$ classes.
RGB and thermal images are accompanied by pixel-wise semantic label maps, while LiDAR and radar measurements are accompanied by point-wise semantic labels.
The complete semantic taxonomy and representative examples are shown in \cref{fig_multifly_sample},
and the dataset is organized hierarchically by scene, acquisition altitude, modality, and synchronized sample.
Calibration parameters, including intrinsic and extrinsic parameters,
as well as synchronization metadata and the complete navigation stream are provided alongside the perception data.

\begin{table}[t]
\centering
\caption{Modal-wise label transfer evaluation (Sec. \ref{sec_eval}).}
\label{tab_label_transfer_eval}
\setlength{\aboverulesep}{0pt}
\setlength{\belowrulesep}{0pt}
\renewcommand{\arraystretch}{1.1}
\setlength{\extrarowheight}{1pt}
\resizebox{\linewidth}{!}{%
\begin{tabular}{l|cccc|c}
\toprule
Altitude & RGB & Thermal & LiDAR & Radar & Average \\
\midrule
\SI{30}{m} & \SI{91.78}{\%} & \SI{90.51}{\%} & \SI{91.80}{\%} & \SI{85.92}{\%} & {\bfseries\SI{90.00}{\%}} \\
\SI{50}{m} & \SI{92.47}{\%} & \SI{90.11}{\%} & \SI{91.11}{\%} & \SI{85.76}{\%} & {\bfseries\SI{89.86}{\%}} \\
\midrule
Average & {\bfseries\SI{92.12}{\%}} & {\bfseries\SI{90.31}{\%}} & {\bfseries\SI{91.45}{\%}} & {\bfseries\SI{85.84}{\%}} & \colorbox[HTML]{B8FFB8}{\bfseries\SI{89.93}{\%}} \\
\bottomrule
\end{tabular}%
}
\end{table}

\section{Dataset Evaluation}
\label{sec_eval}

\subsection{Modal-Wise Label Transfer Evaluation}
\label{subsec_label_transfer_evaluation}

\textbf{RGB \& Thermal.}\quad
We follow SegFly's RGB--thermal evaluation protocol~\cite{gross2026segfly} on MultiFly scene \#4.
The generated pseudo-labels are evaluated against $40$ frames ($10$ for RGB and $10$ for thermal at both altitudes) that were manually annotated solely for evaluation and held out from all stages of data generation (annotation cost: \SI{29}{min} on average per frame).
Results are quantified in \cref{tab_label_transfer_eval}.
On average, MultiFly achieves \SI{92.12}{\%} accuracy for RGB and \SI{90.31}{\%} for thermal, on par with SegFly (\SI{91.27}{\%} / \SI{87.65}{\%}).

\textbf{LiDAR \& Radar.}\quad
We evaluate the transferred 3D labels by rendering
the annotated LiDAR and radar scans
into the same held-out RGB labels using the identical visibility-aware Z-buffered procedure employed during data generation.
Agreement with the held-out manual annotations therefore measures both semantic label quality and geometric consistency of the LiDAR/radar point clouds with the RGB camera.
MultiFly achieves high-quality label transfer accuracy with $91.45$\% for LiDAR and $85.84$\% for radar.

Across all four modalities, the average accuracy is $89.93$\%, confirming high semantic and geometric fidelity of our annotation-efficient label transfer.

\ADLinactivate
\begin{table}[t]
\centering
\caption{Cross-modal semantic consistency across all modalities and scenes of MultiFly
(see \cref{subsec_cross_modal_semantic_consistency}).}
\label{tab_cmsc}

\small
\renewcommand{\arraystretch}{1.1}

\resizebox{\linewidth}{!}{%
\begin{tabular}{l|cccc|c}
\toprule
\multirow{2}{*}{Modality Pair}
    & \multicolumn{4}{c|}{MultiFly Scene}
    & \multirow{2}{*}{Average} \\
\cmidrule(lr){2-5}
    & \#1 & \#2 & \#3 & \#4 & \\
\midrule

\rowcolor[HTML]{E6E6E6}
\multicolumn{6}{l}{\textbf{2D--2D}} \\
\rowcolor[HTML]{FFFFFF}
\hline
RGB--Thermal
    & \SI{90.82}{\%} &\SI{89.83}{\%} &\SI{91.97}{\%} & \SI{91.47}{\%} & {\bfseries\SI{91.02}{\%}} \\

\hline
\rowcolor[HTML]{E6E6E6}
\multicolumn{6}{l}{\textbf{2D--3D}} \\
\rowcolor[HTML]{FFFFFF}
\hline
RGB--LiDAR
   & \SI{91.09}{\%} & \SI{91.41}{\%} & \SI{90.96}{\%} & \SI{92.79}{\%} & {\bfseries\SI{91.56}{\%}} \\
RGB--Radar
   & \SI{91.89}{\%} & \SI{89.14}{\%} & \SI{85.84}{\%} & \SI{88.07}{\%} & {\bfseries\SI{88.73}{\%}} \\
Thermal--LiDAR
   & \SI{91.37}{\%} & \SI{91.79}{\%} & \SI{91.21}{\%} & \SI{93.15}{\%} & {\bfseries\SI{91.88}{\%}} \\
Thermal--Radar
   & \SI{92.16}{\%} & \SI{89.52}{\%} & \SI{86.11}{\%} & \SI{88.44}{\%} & {\bfseries\SI{89.06}{\%}} \\

\hline
\rowcolor[HTML]{E6E6E6}
\multicolumn{6}{l}{\textbf{3D--3D}} \\
\rowcolor[HTML]{FFFFFF}
\hline
LiDAR--Radar
    & \SI{95.40}{\%} & \SI{93.16}{\%} & \SI{92.55}{\%} & \SI{92.58}{\%} & {\bfseries\SI{93.42}{\%}}  \\

\midrule
\textbf{Average}
    & {\bfseries\SI{92.12}{\%}} & {\bfseries\SI{90.81}{\%}} & {\bfseries\SI{89.77}{\%}} & {\bfseries\SI{91.08}{\%}} & \colorbox[HTML]{B8FFB8}{\bfseries\SI{90.94}{\%}} \\
\bottomrule
\end{tabular}
}
\end{table}
\ADLactivate

\begin{figure}[t]
    \centering
    \includegraphics[width=\columnwidth]{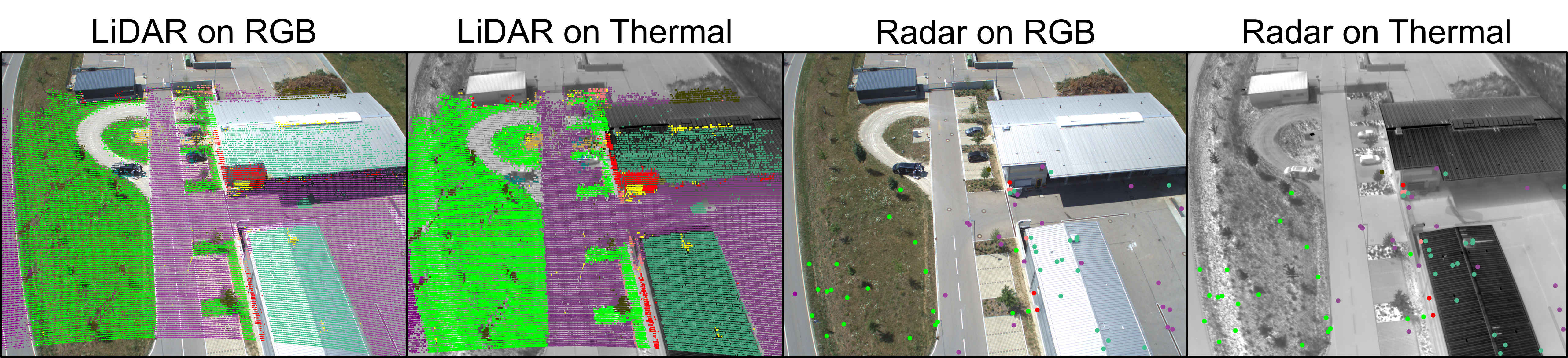}
    \caption{Cross-modal semantic consistency (Sec.~\ref{subsec_cross_modal_semantic_consistency}).}
    \label{fig_semantic_consistency}
\end{figure}

\subsection{Cross-Modal Semantic Consistency Evaluation}
\label{subsec_cross_modal_semantic_consistency}

\textbf{Evaluation Protocol.}\quad
We evaluate the semantic consistency of the final MultiFly labels across all modality pairs without relying on intermediate 3D representations or additional manual ground truth.
Results are summarized in \cref{tab_cmsc}.
We distinguish three types of modality pairs:

\textit{2D--2D (RGB--Thermal)}:
We reuse the geometry-driven 2D--3D--2D registration (\cref{subsec_registration}), but apply it to the RGB semantic labels instead of the RGB image.
The labels are warped onto the thermal image grid via the recovered mapping, after which pixel-wise agreement with the thermal labels is measured.

\textit{2D--3D (RGB/Thermal -- LiDAR/Radar)}:
For each synchronized frame, we project the LiDAR or radar points into the corresponding RGB or thermal image using the calibrated extrinsics, intrinsics.
The semantic label of each valid projected point is compared with the label of the projected pixel.

\textit{3D--3D (LiDAR--Radar)}:
Since radar returns are substantially sparser than LiDAR point clouds, we establish cross-modal correspondences using nearest-neighbor association. For each radar point, we identify the closest LiDAR point within a fixed Euclidean distance threshold $r$ and compare their semantic labels. We set $r=1\,\mathrm{m}$, which provides a correspondence for \SI{71}{\%} of the radar points. Semantic agreement is then determined over these points.
We further evaluate the sensitivity to the choice of $r$ in the results.

For all modality pairs, only pixels/points that yield a valid semantic class (i.e., neither \(c_{\mathrm{unl}}\) nor \(c_{\mathrm{inv}}\)) are considered.
We report consistency as the percentage of agreeing label pairs.

\textbf{Results.}\quad
As shown in \cref{tab_cmsc}, MultiFly achieves an average cross-modal semantic consistency of \SI{90.94}{\%} across all six modality pairs and all four scenes.
For the LiDAR--Radar pair, the consistency remains robust to the choice of association radius, with $r=\{0.1, 0.25, 0.5, 1.0, 2.0\}\,\mathrm{m}$ yielding $\{96, 95, 95, 93, 93\}\,\%$ agreement, respectively, while an unconstrained nearest-neighbor association ($r=\infty$) yields $86\,\%$.
Since all evaluations rely on the calibrated intrinsics, extrinsics, and the RGB--thermal registration, the high agreement also serves as a proxy for geometric fidelity.

Notably, this performance is achieved although our approach employs static 3D reconstruction for label propagation and point-cloud aggregation, which could affect dynamic objects.
While 4D reconstruction methods~\cite{qian2026flow4r} are a promising direction for addressing this limitation in future work, our results confirm that our label transfer ensures both semantically consistent and geometrically coherent multimodal aerial data under this challenging setting.

\subsection{Annotation Efficiency}
\label{subsec_annotation_efficiency}

Only $115$ ($0.67\%$) of MultiFly's $17{,}272$ RGB frames serve as manually annotated source views, taking \SI{29}{min} on average per frame.
From this small subset, the geometry-driven pipeline automatically
labels the remaining $17,157$ RGB frames, all $17,272$ thermal frames, $8.4 \times 10^8$ LiDAR points, and $3.4 \times 10^6$ radar points.
No thermal frame or 3D point is manually annotated for data generation.
The resulting labels exhibit high semantic quality within each modality (\cref{tab_label_transfer_eval}), and strong semantic consistency across modalities (\cref{tab_cmsc}).
Taken together, these results provide strong empirical evidence that sparse RGB annotations can be automatically propagated at scale to generate high-quality, semantically consistent 2D and 3D labels across four modalities without manual refinement.

\section{Benchmark Experiments}
\label{sec_benchmarks}

\ADLinactivate
\begin{table}[t]
\centering
\caption{RGB and Thermal experiments
(\cref{sec_rgb_and_thermal_semantic_segmentation}).}
\label{tab_rgb_thermal_experiments}

\setlength{\aboverulesep}{0pt}
\setlength{\belowrulesep}{0pt}
\renewcommand{\arraystretch}{1.0}
\setlength{\extrarowheight}{1pt}

\resizebox{\linewidth}{!}{%
\begin{tabular}{l|ccc}
\toprule
    & UPerNet~\cite{upernet}
    & SegFormer~\cite{segformer}
    & Firefly~\cite{gross2026segfly} \\
\midrule

\rowcolor[HTML]{E6E6E6}
\multicolumn{4}{l}{\textbf{RGB}} \\
\rowcolor[HTML]{FFFFFF}
\hline
Mean accuracy
    & $53.17\%$
    & $\underline{55.62\%}$
    & $\mathbf{58.21\%}$ \\
Mean intersection over union
    & $39.96\%$
    & $\underline{41.77\%}$
    & $\mathbf{44.05\%}$ \\

\hline
\rowcolor[HTML]{E6E6E6}
\multicolumn{4}{l}{\textbf{Thermal}} \\
\rowcolor[HTML]{FFFFFF}
\hline
Mean accuracy
    & $44.22\%$
    & $\underline{47.81\%}$
    & $\mathbf{53.42\%}$ \\
Mean intersection over union
    & $32.08\%$
    & $\underline{34.71\%}$
    & $\mathbf{40.24\%}$ \\

\bottomrule
\end{tabular}%
}
\end{table}
\ADLactivate

\begin{figure}[t]
    \centering
    \includegraphics[width=\columnwidth]{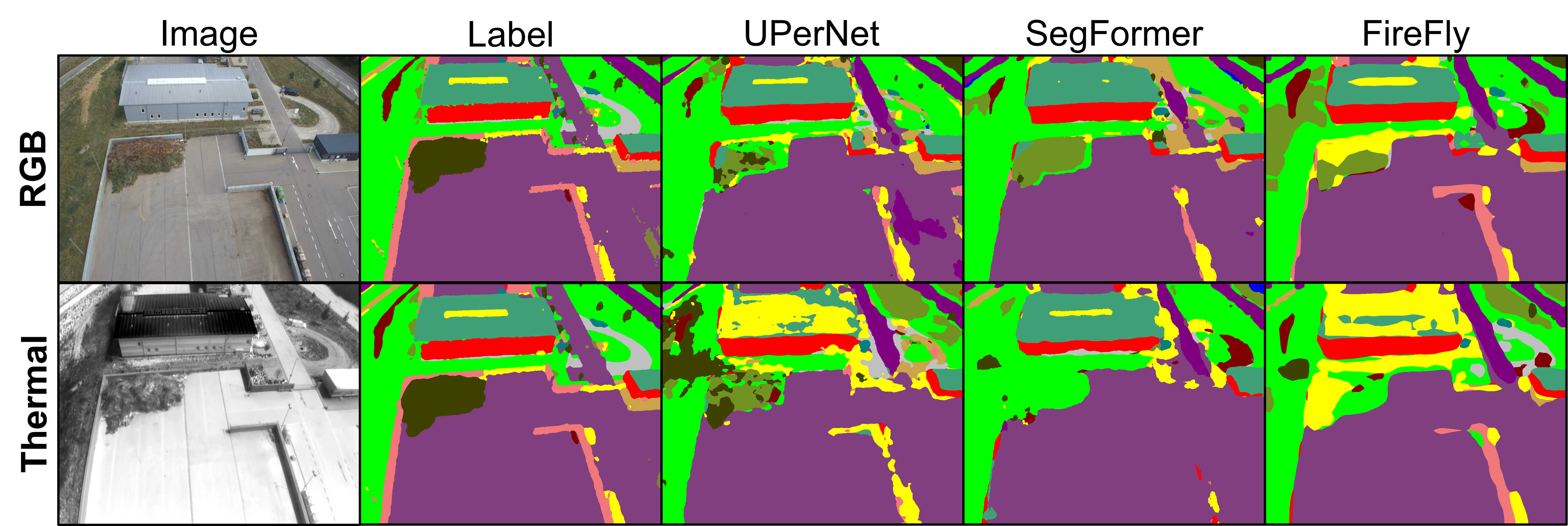}
    \caption{RGB and Thermal experiments (see \cref{sec_rgb_and_thermal_semantic_segmentation}).}
    \label{fig_rgb_thermal_experiments}
\end{figure}

\ADLinactivate
\begin{table}[t]
\centering
\caption{LiDAR and Radar experiments
(\cref{sec_lidar_and_radar_semantic_segmentation}).}
\label{tab_lidar_radar_experiments}

\setlength{\aboverulesep}{0pt}
\setlength{\belowrulesep}{0pt}
\renewcommand{\arraystretch}{1.0}
\setlength{\extrarowheight}{1pt}

\resizebox{\linewidth}{!}{%
\begin{tabular}{l|ccc}
\toprule
    & SpUNet~\cite{pointcept2023spunet}
    & LitePT~\cite{yue2026litept}
    & PTv3~\cite{wu2024ptv3} \\
\midrule

\rowcolor[HTML]{E6E6E6}
\multicolumn{4}{l}{\textbf{LiDAR}} \\
\rowcolor[HTML]{FFFFFF}
\hline
Mean accuracy
    & $49.49\%$
    & $\underline{49.91\%}$
    & $\mathbf{51.76\%}$ \\
Mean intersection over union
    & $34.72\%$
    & $\underline{35.62\%}$
    & $\mathbf{37.09\%}$ \\

\hline
\rowcolor[HTML]{E6E6E6}
\multicolumn{4}{l}{\textbf{Radar}} \\
\rowcolor[HTML]{FFFFFF}
\hline
Mean accuracy
    & $\mathbf{25.95\%}$
    & $21.34\%$
    & $\underline{22.68\%}$ \\
Mean intersection over union
    & $\mathbf{14.50\%}$
    & $12.09\%$
    & $\underline{12.90\%}$ \\

\bottomrule
\end{tabular}%
}
\end{table}
\ADLactivate

\begin{figure}[t]
    \centering
    \includegraphics[width=\columnwidth]{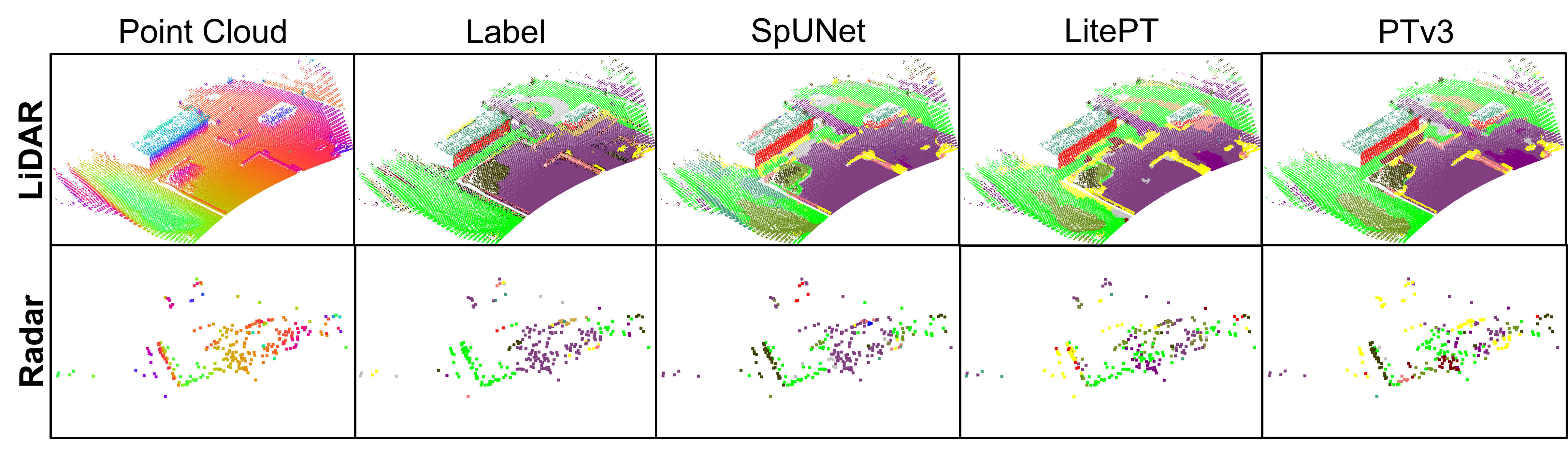}
    \caption{LiDAR and Radar experiments (see \cref{sec_lidar_and_radar_semantic_segmentation}).}
    \label{fig_lidar_radar_experiments}
\end{figure}

\subsection{Experimental Setup}
\label{sec_experimental_setup}

We benchmark established semantic segmentation architectures to assess whether our dataset provides effective supervision for learning semantic representations from its annotated modalities.
The RGB and thermal benchmarks follow the cross-modal training protocol of SegFly~\cite{gross2026segfly}, including RGB-to-thermal adaptation using registered image pairs. LiDAR and radar models are trained independently for their respective modalities. The benchmark therefore evaluates both cross-modal image adaptation and modality-specific 3D segmentation, while dataset-level semantic consistency is assessed separately in \cref{subsec_cross_modal_semantic_consistency}.

Leveraging MultiFly’s four distinct scenes, we adopt a strict held-out-scene protocol to assess scene-level generalization, using scenes~1--3 exclusively for training and scene~4 exclusively for final evaluation.
All models are evaluated using their official configurations.

\subsection{RGB and Thermal Semantic Segmentation}
\label{sec_rgb_and_thermal_semantic_segmentation}

We benchmark established UPerNet-Swin-S~\cite{upernet}, SegFormer-MiT-B3~\cite{segformer}, and Firefly~\cite{gross2026segfly} for RGB and thermal semantic segmentation.
All methods use their official implementations and configurations, including pretrained encoders or backbones, but without external task-specific segmentation checkpoints.
Following SegFly~\cite{gross2026segfly}, training comprises three stages: RGB training, RGB-to-thermal adaptation using registered MultiFly RGB--thermal pairs, and thermal fine-tuning.
All models are trained for 30 epochs for RGB training, 10 epochs for RGB-to-thermal adaptation, and 30 epochs for thermal fine-tuning.
We report the final-epoch results.

As shown in Table~\ref{tab_rgb_thermal_experiments}, Firefly achieves the strongest performance for both RGB and thermal semantic segmentation in terms of mean intersection over union (mIoU) and mean accuracy (mAcc).
Qualitative predictions for a representative sample in \cref{fig_rgb_thermal_experiments} are consistent with the quantitative results, demonstrating that MultiFly supports effective semantic segmentation across both modalities.

\subsection{LiDAR and Radar Semantic Segmentation}
\label{sec_lidar_and_radar_semantic_segmentation}

We benchmark state-of-the-art SparseUNet (SpUNet)~\cite{pointcept2023spunet}, LitePointTransformer-S (LitePT)~\cite{yue2026litept}, and PointTransformerV3 (PTv3)~\cite{wu2024ptv3} independently on the LiDAR and radar labels.
All methods use their official implementations and configurations unless stated otherwise.
In addition to 3D coordinates, the LiDAR inputs include intensity, while the radar inputs include radar cross section (RCS) and Doppler velocity, in line with~\cite{arroyo2024radarUnimodalDataset}.
Each model is trained separately for LiDAR and radar semantic segmentation for 20 epochs, and we report the final-epoch results.
For radar, the only configuration change is an increase in the voxelization grid-size from \SI{0.05}{m} to \SI{0.50}{m}, providing a coarser discretization
to account for the lower density of radar points.

As shown in Table~\ref{tab_lidar_radar_experiments}, PTv3 performs best on LiDAR, whereas SpUNet achieves the strongest radar performance.
Prior work identifies sparse-convolutional approaches for sparser radar point clouds as an understudied area~\cite{arroyo2024radarUnimodalDataset}.
By benchmarking SpUNet against recent point-transformer architectures, we directly address this gap and find that established sparse convolutions can outperform recent attention-based models.
This result provides a promising initial benchmark for radar semantic segmentation, while controlled studies are required to disentangle the effects of individual architectural choices.
Qualitative predictions for a representative sample in \cref{fig_lidar_radar_experiments} further support the quantitative results.

\section{Conclusion}
\label{sec_conclusion}

We introduce MultiFly, a real-world low-altitude UAV dataset with synchronized RGB, thermal, LiDAR, and radar semantic annotations.
By propagating labels from only 115 manually annotated RGB images, MultiFly provides dense supervision across 17{,}272 multimodal samples with high annotation agreement and cross-modal consistency.
Our benchmarks further establish initial semantic segmentation baselines for all four modalities.

\bibliographystyle{IEEEtran}
\bibliography{references.bib}

\end{document}